\documentclass[letterpaper, 10 pt, conference]{ieeeconf}  

\IEEEoverridecommandlockouts                              

\usepackage{graphics} 
\usepackage{epsfig} 
\usepackage{mathptmx} 
\usepackage{times} 
\usepackage{amsmath} 
\usepackage{amssymb}  

\usepackage{url}
\usepackage{booktabs} 

\title{\LARGE \bf
From One to Many: How Active Robot Swarm Sizes \\Influence Human Cognitive Processes
}

\author{Julian Kaduk,$^{1}$ Müge Cavdan,$^{2}$ Knut Drewing$^{2}$ and Heiko Hamann$^{1}$
\thanks{*This work was supported by European Union’s Horizon 2020 FET Open research program, grant no 964464, project ChronoPilot and by Deutsche Forschungsgem., Germany's Excellence Strategy EXC 2117-422037984.}
\thanks{$^{1}$Julian Kaduk and Heiko Hamann are with the Department of Computer and Information Science,
        University of Konstanz, 78457 Konstanz, Germany
        {\tt\small julian.kaduk@uni-konstanz.de},
        {\tt\small heiko.hamann@uni-konstanz.de}}%
\thanks{$^{2}$Müge Cavdan and Knut Drewing are with the Experimental Psychology Department, Justus Liebig University,
        35390 Giessen, Germany
        {\tt\small muege.cavdan@psychol.uni-giessen.de},
        {\tt\small knut.drewing@psychol.uni-giessen.de}}%
}

\begin{document}

\maketitle
\thispagestyle{empty}
\pagestyle{empty}

\begin{abstract}

In robotics, understanding human interaction with autonomous systems is crucial for enhancing collaborative technologies. We focus on human-swarm interaction (HSI), exploring how differently sized groups of active robots affect operators' cognitive and perceptual reactions over different durations. We analyze the impact of different numbers of active robots within a 15-robot swarm on operators' time perception, emotional state, flow experience, and task difficulty perception. Our findings indicate that managing multiple active robots when compared to one active robot significantly alters time perception and flow experience, leading to a faster passage of time and increased flow. More active robots and extended durations cause increased emotional arousal and perceived task difficulty, highlighting the interaction between robot the number of active robots and human cognitive processes. These insights inform the creation of intuitive human-swarm interfaces and aid in developing swarm robotic systems aligned with human cognitive structures, enhancing human-robot collaboration.

\end{abstract}

\section{Introduction}\label{section:intro}   
Human-swarm interaction (HSI) is an emerging field that focuses on the dynamics between human operators and swarms of robotic agents~\cite{Kolling.2015} going beyond human-robot interaction (HRI). As technological advancements enable the deployment of increasingly large and complex swarms, understanding how humans perceive, interpret, and manage these systems becomes critical. The interplay between human cognition and collective robot behavior presents unique challenges and opportunities, particularly in situations where human operators must maintain situational awareness and control over the swarm's actions.

A robot swarm consists of many simple robots that operate autonomously through local interactions without central control. This allows for complex collective behavior to emerge from simple individual behaviors, with the added benefits of scalability, flexibility, and reliability~\cite{Dorigo.2004}. Although most of the literature on swarm robotics focuses on fully autonomous systems~\cite{Hamann.2018}, introducing the human factor and going partially semi-autonomous presents a significant research venture. A~human supervisor should manage the potentially large number of individual robots, which can be difficult to oversee simultaneously. Research has indicated that the amount of operator influence should be usefully limited to improve the overall performance of the swarm~\cite{Ashcraft.2019}, and even the timing of user input is critical to avoid interrupting emerging behaviors~\cite{Nagavalli.2014}. Consequently, a~major focus in HSI is on designing different user interfaces that allow for human control of a swarm without eliminating the swarm's characteristics or compromising mission performance.

Our focus is on a secondary and under-explored aspect of HSI that looks at the effects of the swarm on human perception in an interactive context. We specifically look at the effects of swarm configurations on how humans affectively and cognitively process the control task. Generally, we want to explore nuanced aspects of HSI by examining how variations in the number of active robots in a swarm influence human perception, cognition, and task performance. Here, we study the effects on time perception, emotional responses, flow experience, and perceived task difficulty. These are central aspects of human processing that do not only mediate resulting task performance but also long-term satisfaction with robot collaboration and well-being.

In a previous study, we found that a higher number of robots in a user-controlled swarm resulted in the perception that time passed faster in combination with an effect on the flow experience of the participants~\cite{Kaduk.2023}. We see the opportunity of designing HSI to possibly leverage this perception-modulating effect by adaptively and dynamically disabling and reactivating robots in the swarm to react to human needs. To investigate this approach we designed an experiment by dividing a swarm of~$N=15$ robots into two sub-swarms of moving (active) and unmoved (passive) robots with different swarm splits for each trial. With this careful experiment design, we contribute to a more precise understanding of how quantitative changes in active robots affect qualitative aspects of the human experience.

Through a series of experiments, we examine the participants' ability to manage sub-swarms of varying sizes over different durations, assessing their time perception, emotional state, flow experience, and cognitive load. Based on our results, we hope to improve the future development of more effective human-swarm interfaces and to provide guidelines for optimizing swarm behavior to align with human cognitive processes when interacting with configurations of active and passive robots. By advancing our understanding of HSI, we pave the way for more intuitive and efficient collaborative systems where humans and groups of robots can achieve shared goals with enhanced synergy and mutual understanding.

\section{Related work}\label{section:literature} 
Research in HSI is predominantly focused on optimizing interfaces and control mechanisms to facilitate effective human oversight while leveraging the advanced capabilities of swarm intelligence. This balance aims to ensure that the swarm's autonomy is not undermined by human intervention, yet the system benefits from human superior cognitive abilities for strategic oversight~\cite{Kolling.2016}. It has for example been shown as advantageous to provide the operator with a global overview of the swarm-state through a heat-map representation rather than presenting data from each agent~\cite{Divband.2021}. The use of augmented reality~\cite{millard.2018} and virtual reality~\cite{jang2021} has also been shown to be effective tools in HSI.

Beyond the means of interaction, it is crucial to consider how the swarm control affects the user's affective and cognitive processing of the task. In particular, trust within the context of HSI is identified as a key factor that influences the effectiveness of human-swarm collaboration. Enhancing the operator's trust involves providing them with accurate and timely feedback about the swarm's state, thereby improving the decision-making process and task performance~\cite{Abioye.2023}. The challenge lies in designing interfaces that provide simplified, but informative feedback to operators, enabling them to make informed decisions without being overwhelmed by the swarm's complexities~\cite{Hussein.2018}. A~possible holistic vision is a ``joint human–swarm loop" in which the human becomes a part of the swarm and the robot an extension of the human~\cite{Hasbach.2022}. 

Fully integrating the human within the swarm requires an understanding of how the different characteristics in the swarm behavior affect the affective and cognitive processing of a human operator. Research has shown that in an interaction scenario where the human remains passive (only watching a swarm), a higher number of robots results in increased self-reported levels of arousal~\cite{Podevijn.20164ow}. Higher robot speeds cause increased reported arousal levels and smoother robot motion causes higher levels of reported emotional valence~\cite{Dietz.2017}. In our previous work, we identified similar effects in an active interaction scenario of human participants with a robot swarm and focused on the additional variable of how time perception and flow experience can be influenced, showing that more robots in a swarm and an increased speed can cause an increased flow experience and sped up time perception~\cite{Kaduk.2023}.

Here, we focus on how time perception, emotional state, flow experience, and cognitive load are affected by changing the number of active robots within the swarm while keeping the total swarm size constant with~$N=15$ robots. These human factors are fundamental aspects of our affective and cognitive processes, and to our knowledge, no study has considered them in the given context. In the following, we will explain the aforementioned aspects and their relevance in greater detail. 

Time is ubiquitous - any sensory experience, irrespective of its modality, contains a temporal component. Therefore, in executing daily tasks, we often rely on our perception of time to effortlessly guide our actions. However, time perception is a subjective experience and is susceptible to distortions. This subjective experience changes depending on various cognitive, perceptual, and emotional factors~\cite{Wittmann.2015}. For example, the crowding level inside a subway car impacts the time perception of a journey, with more passengers leading to lengthened time~\cite{Sadeghi.2023}. 

The objective factors that influence our perception of time perception vary, such as size~\cite{Thomas.1975}, numerosity~\cite{Dormal.2006}, motion~\cite{Brown.1995}, or the emotional value of a stimulus~\cite{Noulhiane.2007, Angrilli.1997}. For example, rapidly moving stimuli lengthened the perception of time compared to slower or stationary stimuli. In our daily experiences, we rarely encounter a visual scene that contains exclusively static or dynamic objects but rather consists of a mixture of both. In an active robot supervision task, this presents additional cognitive complexity, as all robots are visually very similar. To differentiate which targets are relevant for the supervision task, one first has to mentally separate them from the rest. Vision research has shown that moving targets are easier for humans to detect than stationary targets~\cite{matsuno2006visual}. However, with multiple relevant targets, the visual search likely has to be repeated continuously if the participant is not able to keep track of all moving agents simultaneously. 

The emotional state is not only highly related to our perception of time~\cite{Wittmann_2009} but also describes our well-being and can even be closely linked with cognitive load~\cite{plass.2019}. As we strive for high levels of well-being in the interaction with the robots, it is therefore crucial to consider these factors with the perceived task difficulty being our measure of the cognitive load. Furthermore, flow represents an optimal state of cognitive involvement, situated between boredom and anxiety, characterized by a harmonious balance between the skills of an individual and the demands of the task at hand~\cite{csikszentmihalyi1990flow}. The experience of flow is also frequently associated with altered perceptions of time~\cite{flow_time} and considered as a desirable state of high well-being.

\section{Method}\label{section:method}

\subsection{Scenario}
\begin{figure}[t]
    \vspace{8pt}
    \centering
    \includegraphics[height=4.8cm]{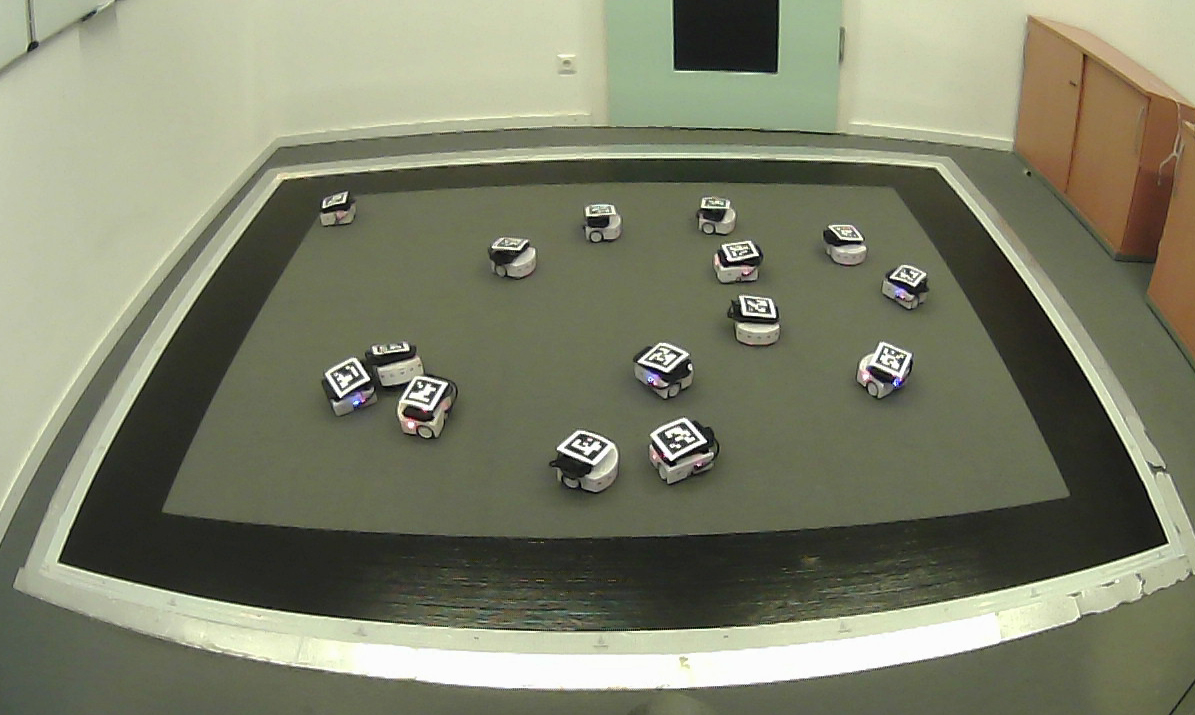}
    \caption{View of the arena from the perspective of a participant.}
    \label{fig:participant}    
\end{figure}

The task of our participants was to prevent a sub-swarm of small mobile robots from exiting a 2.2~m by 1.6~m robot arena, delineated with black tape on the ground. While the robots exhibited a random walk behavior, described in detail in Sec.~\ref{subsec:behavior}, participants were seated next to the arena and provided with a single-button interface to try to supervise the robots' movements. The perspective of the participant is shown in Fig.~\ref{fig:participant}.

We tested eight different sub-swarm sizes $N_{\text{a}} \in \{1, 3, 5, 7, 9, 11, 13, 15\}$, with $N_{\text{a}}$ being the count of active robots from a total of $N=15$ in the swarm. 
The $N_{\text{p}}=N-N_{\text{a}}$ passive robots were placed and remained stationary in random locations within the arena throughout the trial. 
The experiment was carried out for three distinct durations $t \in \{1, 3, 5\}$ given in minutes, totaling 24 unique trial conditions. 
These were randomly presented to participants to minimize order effects across two sessions, each lasting approximately 1.5~hours, a structure chosen to reduce participant fatigue.

\subsection{Experiment Procedure}
The initial session began with a briefing on the experiment and the overall research project, followed by the participants signing an informed consent form and completing an anonymous demographic questionnaire. Subsequently, participants engaged in several randomized trials. Before each trial, they watched a relaxing aquatic video for two minutes to establish a baseline condition~\cite{Piferi_Kline_Younger_Lawler_2000}. This preceded the main trial, which involved interacting with robots for durations of one, three, or five minutes, and ended with the participants filling out a six-item post-trial questionnaire to assess their time perception and experience. Participants had been introduced to the questionnaire before the first trial to avoid initial bias.

Ethical approval for this study was obtained from the Ethics Committee of the University of Konstanz\footnote{reference number: 24/2023} and was in accordance with the Declaration of Helsinki~\cite{Helsinki.2013} without pre-registration.

\subsection{Robot Hardware and Behavior}\label{subsec:behavior}
We used Thymio II robots~\cite{Riedo_Chevalier_Magnenat_Mondada_2013}, which were augmented with a Raspberry~Pi~3B to enable wireless connectivity and control through a Python interface, thus improving their capabilities. The Thymio~II robot, with dimensions of 110~mm by 112~mm, operates on a differential drive system with two motors and can reach a maximum linear velocity of 0.2~$\frac{m}{s}$. It is equipped with seven infrared sensors mounted horizontally, five at the front and two at the rear, each with a detection range of about 100~mm.

The robot behavior was programmed to integrate a random walk with obstacle avoidance and user input. During the random walk, the robot proceeds straight at a constant speed of $v = 0.2 \frac{m}{s}$ for a randomly chosen duration between one and ten seconds, based on a uniform distribution. If an obstacle is detected by its sensors or after the random duration elapses, the robot rotates in place in a random direction at a constant angular velocity for a random period between 0.5 and 3~seconds, determined by a uniform distribution. Subsequently, it resumes moving straight, selecting a new random duration.

The user interaction was facilitated through a button press, triggering all robots to spin in place and alter their direction by inverting the velocity of one wheel until the button was released or a button timeout of three seconds had elapsed. This control mechanism allowed participants to manage the robots' movements within the designated area. To prevent overuse of the button, a three-second timeout was implemented, which resets each time the button is released or after the timeout period, allowing the robots to continue their programmed behavior.

\subsection{Self-reported Measures}
After each trial, participants were asked to complete a survey comprising six questions distributed in three categories. The first category aimed at evaluating their time perception, encompassing questions on estimating the interaction duration and assessing the speed of time passage. Participants estimated the interaction length on a visual timeline from 0:00 to 10:00 minutes and rated the perceived speed of time on a 5-point Likert scale, ranging from `very slow' to `very fast'. The second category assessed the participants' affective states with two questions about emotional valence and arousal, using the 9-point Likert scale of the Self-Assessment Manikin~\cite{Bradley_Lang_1994}. The final category asked participants about their experience with the flow state and perceived difficulty of the task, utilizing 9-point and 5-point Likert scales, respectively, to measure their level of involvement and challenge of the task.

\subsection{Participants}
We recruited 18 participants for this study with 50\% identifying as male, 50\% as female. Five participants were in the age range of 18 to 20 years and 13 participants from 21 to 29 years. The eligibility criteria included a minimum age of 18 years, proficient written comprehension of the German language for the questionnaires, and the absence of any known current or prior psychological disorders or cardiac diseases. No financial compensation was provided for participation in this study.

\section{Results}\label{section:results}  
Statistical analyses were conducted in R Studio 2023 version 4.3.0 using Linear Mixed Models (LMM). The LMM's were fitted with the \textit{lmer} library by the maximum likelihood estimation. For every dependent variable (i.e., time estimation, passage of time, flow, arousal, valence, and difficulty), each participant was entered into the model as a random intercept to accommodate variations between individuals, allowing for more accurate estimates of the fixed effects. For every model, the fixed effects were the number of active robots, duration, and interaction between the number of robots and duration. The fixed effect results were estimated with the \textit{anova} function, which calculates an \textit{F~test} on the fixed effects using the Satterthwaite approximation. Follow-up post-hoc pairwise comparisons and means were estimated using the \textit{emmeans} library. The Bonferroni adjustment method was used for multiple pairwise comparisons if needed.

 \begin{table}[t]
 \vspace{8pt}
 \scriptsize
   \caption{Fixed effects (Number of Robots, experiment duration, interaction) in LMMs for time estimation, passage of time, flow, task difficulty, arousal, and valence, reported with sum of squares (SS), mean squares (MS), degrees of freedom (df), F-value (F), and p-value (p).}
   
   
   \label{tab:results}
   \centering
   \begin{tabular}{l l l l l l}
     Fixed effect & SS & MS & df & F & p \\
     \toprule
     \toprule
     \multicolumn{6}{c}{Passage of time} \\
     \midrule
     Number of Robots  	& 56.98 		& 8.14 		& 7, 414 	& 12.31 		& $<$ 0.001 \\
     Duration 			& 136.06 	& 68.03 		& 2, 414 	& 102.90 	& $<$ 0.001 \\
     NoR $\times$ Duration 	& 8.79 		& 0.63 		& 14, 414 	& 0.95 		& 0.5 \\
     \toprule
     \multicolumn{6}{c}{Time Estimation} \\
     \midrule
     Number of Robots 	& 1.37 		& 0.20 		& 7, 414 	& 0.32 		& 0.98 \\
     Duration 			& 698.20 	& 349.10 	& 2, 414 	& 403.16 	& $<$ 0.001 \\
     NoR $\times$ Duration 	& 3.47 		& 0.25 		& 14, 414 	& 0.29 		& 1.00 \\
     \toprule
     \multicolumn{6}{c}{Flow} \\
     \midrule
     Number of Robots 	& 68.88 		& 9.84 		& 7, 414 	& 3.90 		& $<$ 0.001 \\
     Duration 			& 3.67 		& 1.84 		& 2, 414 	& 0.73 		& 0.48 \\
     NoR $\times$ Duration 	& 21.77 		& 1.56 		& 14, 414 	& 0.62 		& 0.85 \\
     \toprule
     \multicolumn{6}{c}{Perceived task difficulty} \\
     \midrule
     Number of Robots 	& 209.15 	& 29.88 		& 7, 414 	& 57.38 		& $<$ 0.001 \\
     Duration 			& 10.29 		& 5.15 		& 2, 414 	& 9.88 		& $<$ 0.001 \\
     NoR $\times$ Duration 	& 10.26 		& 0.73 		& 14, 414 	& 1.41 		& 0.15 \\
     \toprule
     \multicolumn{6}{c}{Emotional arousal} \\
     \midrule
     Number of Robots 	& 577.80 	& 82.55 		& 7, 414 	& 32.40 		& $<$ 0.01 \\
     Duration 			& 26.48  	& 13.24 		& 2, 414 	& 5.20 		& $<$ 0.001 \\
     NoR $\times$ Duration 	& 36.34 		& 2.60 		& 14, 414 	& 1.02 		& 0.43 \\
     \toprule
     \multicolumn{6}{c}{Emotional valence} \\
     \midrule
     Number of Robots 	& 18.98 		& 2.71 		& 7, 414 	& 1.77 		& 0.09 \\
     Duration 			& 14.69 		& 7.34 		& 2, 414 	& 4.78 		& $<$ 0.01 \\
     NoR $\times$ Duration 	& 33.72 		& 2.42 		& 14, 414 	& 1.57 		& 0.08 \\

   \bottomrule
   \end{tabular}
 \end{table}

\subsection{Passage of time}
The LMM analysis on the effect of duration and number of robots on the passage of time showed the main effects of the number of active robots and duration on the passage of time while there was no interaction effect (see Tab.~\ref{tab:results}). Post-hoc pairwise comparisons revealed that 1-min passed faster compared to 3-min, and 5-min and 3-min passed faster compared to 5-min. Also, pairwise comparisons showed that the 1-robot condition compared to the rest of the number of active robot conditions resulted in a slower passage of time while there was no significant difference between other comparisons (p-value adjustment for 28~tests). The data for this measure are shown in Fig.~\ref{fig:time}.

\subsection{Time estimation}
\begin{figure}[t]
    \centering
    \includegraphics[height=4.3cm]{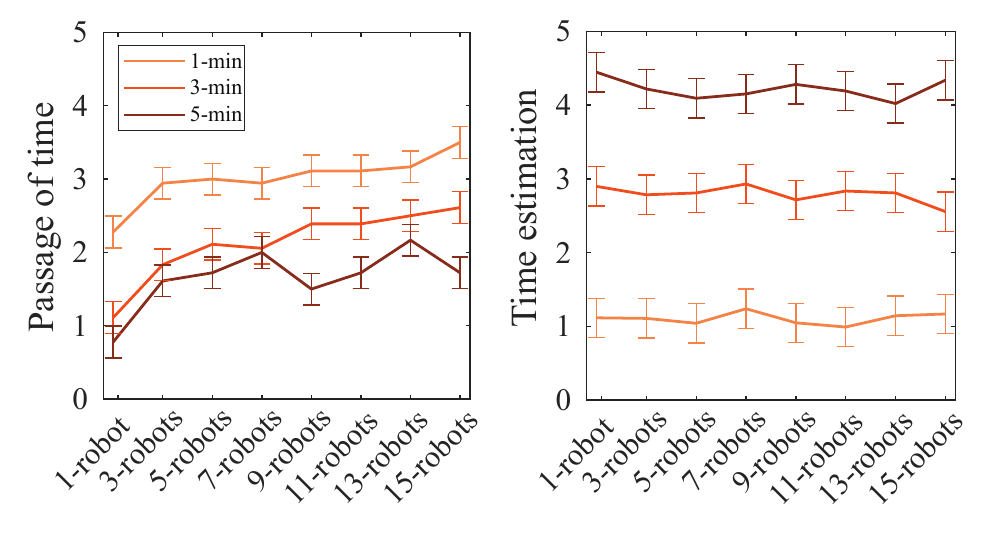}
    \caption{Estimated mean passage of time (left) and time estimation (right) as a function of number of active robots for the self-reported measures on a scale from 0 to 4 for the passage of time and the absolute duration estimation in minutes. Error bars correspond to the standard error of the mean.}
    \label{fig:time}    
\end{figure}

The LMM analysis on the effect of duration and number of active robots on time estimation showed a main effect of duration but not for the number of active robots, or the interaction between the number of active robots and duration (see Tab.~\ref{tab:results} for the detailed results). Follow-up on the duration, expectedly, 5-min and 3-min estimated longer compared to 1-min and 3-min estimated shorter than 5-min (all $p < .001$). The data for this measure are shown in Fig.~\ref{fig:time}.

\subsection{Flow}
The LMM analysis on the effect of duration and number of robots on flow showed only the significant main effect of the number of active robots, but not the effect of duration and their interaction (see Tab.~\ref{tab:results}). Post-hoc pairwise comparisons revealed that 1-robot compared to 5-robot, 7-robot, 9-robot, 13-robot, and 15-robot had lower flow while there was no other significant difference between other comparisons (p-value adjustment for 28~tests). The data for this measure are shown in Fig.~\ref{fig:flow}.

\subsection{Difficulty}
\begin{figure}[t]
    \centering
    \includegraphics[height=4.3cm]{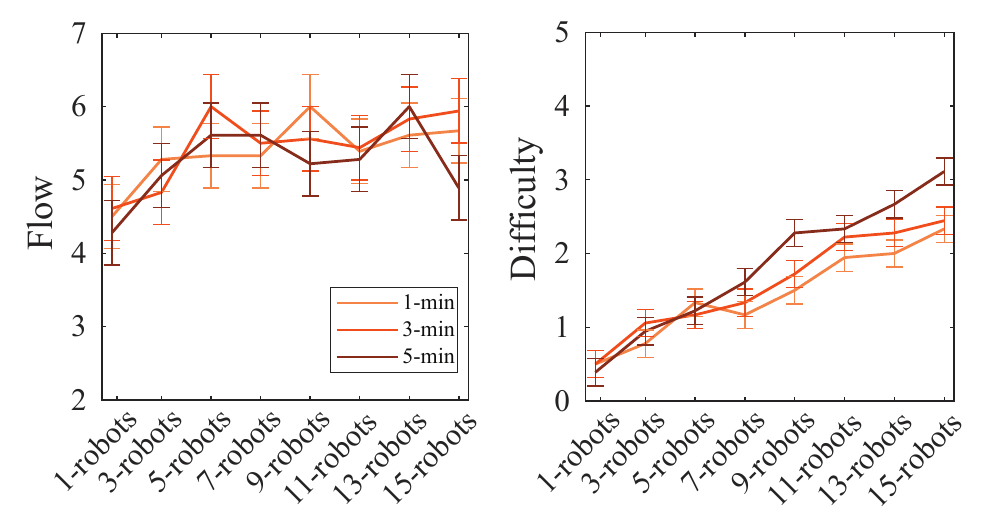}
    \caption{Estimated mean flow (left) and difficulty (right) as a function of number of active robots for the self-reported measures on a scale from 0 to 8 for flow and from 0 to 4 for difficulty. Error bars correspond to the standard error of the mean.}
    \label{fig:flow}    
\end{figure}

The LMM analysis on the effect of duration and number of robots on difficulty revealed the main effect of number of active robots and duration while their interaction was not significant (see Tab.~\ref{tab:results}). Duration follow-up showed that 5-min felt more difficult than 1-min and 3-min (correction for three tests). Also, with a few exceptions (3-robot vs 5-robot and 7-robot, 5-robot vs 7-robot, 9-robot vs 11-robot, 11-robot vs 13-robot, 13-robot vs 15-robot), the increasing number of robots is perceived to be more difficult (p-value correction for 28~tests). The data for this measure are shown in Fig.~\ref{fig:flow}.

\subsection{Arousal}
The LMM analysis on the effect of duration, number of robots, and their interaction on arousal revealed significant main effects of number of robots and duration while their interaction was not significant (see Tab.~\ref{tab:results}). The pairwise comparison of duration showed that three-min and five-min were more arousing compared to 1-min (p-value corrected for three tests). Follow-up results on the number of robots revealed high arousal with the increasing number of robots: 15-robot felt more arousing than others (except 11-robot and 13-robot) which is followed by 13-robot (except 9-robot and 11-robot). 9-robot felt more arousing than 1-robot and 3-robot as well as 7-robot, 5-robot, and 3-robot felt more arousing than 1-robot (p-value adjustment for 28~tests). The data for this measure are shown in Fig.~\ref{fig:emotion}.

\subsection{Valence}
\begin{figure}[t]
    \centering
    \includegraphics[height=4.3cm]{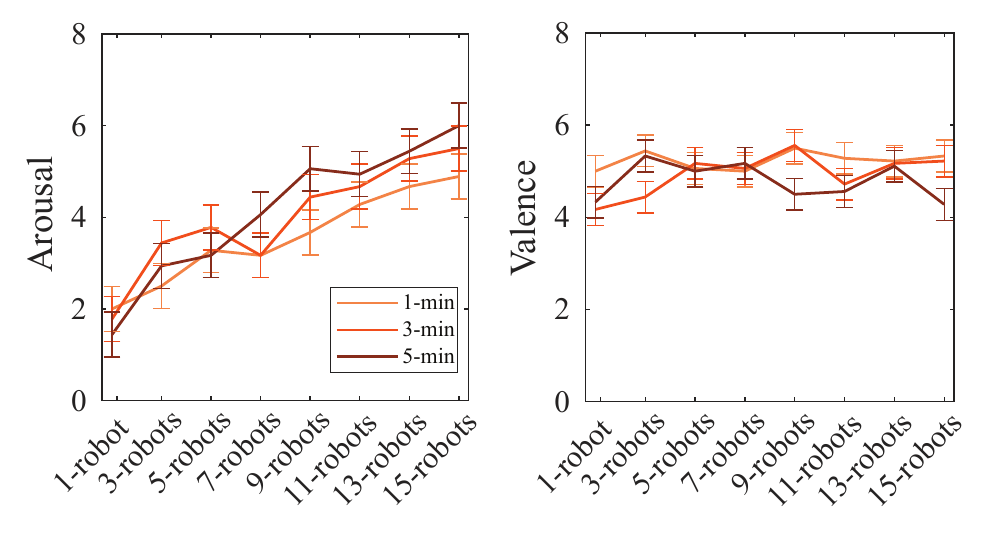}
    \caption{Estimated mean arousal (left) and valence (right) as a function of number of active robots for the self-reported measures on a scale from 0 to 8. Error bars correspond to the standard error of the mean.}
    \label{fig:emotion}    
\end{figure}

The LMM analysis on the effect of duration, number of active robots, and their interaction showed the significant main effect of duration (see Tab.~\ref{tab:results}) while the number of robots and interaction were not statistically meaningful. Follow-up results show that 1-min felt more pleasant than 5-min (p-value corrected for three tests). The data for this measure are shown in Fig.~\ref{fig:emotion}.

\section{Discussion}\label{section:discussion}  
In our investigation, we focused on the impact of different sub-swarm sizes on time perception. Our previous study~\cite{Kaduk.2023} suggested that a larger number of robots when interacting with a complete swarm of active robots only, accelerates perceived time and shortens duration judgments. Here, the human operator interacts only with a subset of active robots and we observed different effects. 

Our findings indicate that the passage of time perception is relatively stable across varying numbers of active robots in all multi-robot conditions. However, in comparison, the single-robot condition resulted in a significantly slower perceived passage of time, suggesting that the mere visual presence of additional inactive robots impacts human time perception. This is a critical difference showing that one cannot simply deactivate robots to adaptively modulate time perception in HSI. We suspect that in the 1-robot condition, the participant can easily follow the robot with their attention while the multi-robot conditions require an additional step of having to identify the relevant, moving robots before being able to decide if a user input is required. 

Regarding the duration of the experiment, we observed that participants tended to underestimate the duration of 5-minute experiments and overestimate 1-minute experiments. This aligns with the fundamental theory of ``Vierodt's law" in time perception, stating that shorter durations tend to be overestimated, while long durations are underestimated~\cite{Lejeune.2009}. While this phenomenon is understood for durations in the sub-second range (e.g.,~\cite{Cavdan.2023a}), our data highlight that it can also be observed for longer durations with an ``indifference point" at somewhere between one and three minutes in our experiment. 

The data on flow experience aligns with the pattern observed in time perception. Participants reported lower flow states in the single-robot condition than in multi-robot conditions, reinforcing the notion that an enhanced perception of being in a state of flow and a faster perception of time are interconnected. However, the duration of the experiment did not influence the flow state, indicating that primarily the number of active robots drives this aspect of user experience rather than the length of interaction. This correlation between an increased sense of flow and larger active sub-swarms highlights the importance of engaging the operator effectively in the task, which seems to be facilitated by the complexity and demands of managing multiple robots.

Our results demonstrate a clear correlation between the number of active robots and the perceived task difficulty, particularly in longer interactions. Larger active sub-swarms were associated with increased cognitive load, potentially due to the higher demand for attention and control required to manage multiple robots. This effect was particularly pronounced in the longest-duration conditions, suggesting that operator fatigue and the associated cognitive load could be significant factors in designing swarm interaction tasks, especially for extended periods.

Considering the emotional effects of different sub-swarm sizes, we find no impact on participants' emotional valence with a neutral average rating. Only the 1-minute condition was overall rated as more pleasant than the 5-minute condition. However, the emotional arousal was significantly heightened with an increase in the number of active robots and longer interaction durations. This heightened arousal could be attributed to the increased engagement and cognitive demand required in these scenarios, indicating that the design of swarm robotic systems needs to consider the potential for inducing stress or over-stimulation in human operators.

These results show that the time perception and flow experience cannot be actively controlled by deactivating parts of the swarm to limit the interaction to an active subset of robots. However, the perceived task difficulty and arousal can be adaptively modulated in this way. As both factors are also negatively affected by longer durations with increased levels of arousal and higher perceived task difficulty, a reduction of robots over time could also be used to balance the cognitive demand on the operator. These insights have several implications for the design of human-swarm interfaces and the overall management of swarm robotic systems. Designers must carefully manage the engagement of operators with active robots as this is vital for creating effective, efficient, and user-friendly swarm robotic systems.

\section{Conclusion}\label{section:conclusion}  
Our study provides significant insights into HSI, highlighting how variations in active sub-swarm sizes impact human cognitive and perceptual responses. We have uncovered that different numbers of active robots distinctly influence time perception, flow experience, emotional arousal, and perceived task difficulty. 
The contrast in our results between scenarios with single and multiple active robots highlights the complexity of human responses to swarm dynamics. 
While adjustments in active sub-swarm sizes do not uniformly alter all human perceptions, they can strategically influence factors such as task difficulty and emotional arousal. These findings are vital for designing intuitive human-swarm interfaces, suggesting a need for adaptive systems that align with human cognitive demands to enhance collaborative efficiency and user satisfaction. 
Our research hopefully lays the foundation for more research on personalized and adaptive swarm behaviors, targeting seamless integration of human-robot collaborations. It reveals the critical need to incorporate human cognitive and perceptual factors into the design of swarm robotics, enhancing collaborative effectiveness.








\bibliography{references.bib}
\bibliographystyle{IEEEtran}


\end{document}